\documentclass[journal]{IEEEtran}

\ifCLASSINFOpdf
\else
   \usepackage[dvips]{graphicx}
\fi
\usepackage{url}

\usepackage[cmintegrals]{newtxmath}
\usepackage{amsmath}
\usepackage[ruled]{algorithm2e}
\usepackage{times}
\usepackage{epsfig}
\usepackage{graphicx}
\usepackage{subfigure}
\usepackage{amsmath}

\usepackage{amssymb}
\usepackage{multirow}
\usepackage{verbatim}
\usepackage{bm}
\usepackage{framed}
\usepackage{makecell}
\usepackage{booktabs}
\usepackage{multirow}

\begin{document}

\title{Face Transformer for Recognition}

\author{Yaoyao Zhong,
	Weihong Deng
	\thanks{The authors are with the Pattern Recognition and Intelligent System
		Laboratory, School of Artificial Intelligence, Beijing University of Posts and Telecommunications, Beijing 100876, China (e-mail: zhongyaoyao@bupt.edu.cn; whdeng@bupt.edu.cn).}
}

\maketitle

\begin{abstract}
Recently there has been a growing interest in Transformer not only in NLP but also in computer vision. We wonder if transformer can be used in face recognition and whether it is better than CNNs. Therefore, we investigate the performance of Transformer models in face recognition. Considering the original Transformer may neglect the inter-patch information, we modify the patch generation process and make the tokens with sliding patches which overlaps with each others. The models are trained on CASIA-WebFace and MS-Celeb-1M databases, and evaluated on several mainstream benchmarks, including LFW, SLLFW, CALFW, CPLFW, TALFW, CFP-FP, AGEDB and IJB-C databases. We demonstrate that Face Transformer models trained on a large-scale database, MS-Celeb-1M, achieve comparable performance as CNN with similar number of parameters and MACs. To facilitate further researches, Face Transformer models and codes are available at \url{https://github.com/zhongyy/Face-Transformer}.
\end{abstract}


\IEEEpeerreviewmaketitle

\section{Introduction}

\IEEEPARstart{R}{ecently} it seems a popular trend to apply Transformer in different computer vision tasks including image classification~\cite{dosovitskiy2020image}, object detection~\cite{carion2020end}, video processing~\cite{zhou2018end} and so on. Although the inner workings of Transformer is not that clear, researchers come up with idea after idea to apply Transformer in different kinds of ways~\cite{touvron2020training,yuan2021tokens,han2021transformer} because of its strong representation ability. 

Based on large-scale training databases~\cite{guo2016msceleb} and effective loss functions~\cite{SphereFace,Wang2018CosFace,deng2019arcface}, convolution neural networks (CNNs), from VGGNet~\cite{simonyan2014very} to ResNet~\cite{he2016deep}, have achieved great success in face recognition over the past few years~\cite{deng2019arcface}. DeepFace~\cite{taigman2014deepface} first uses a 9-layer CNN in face recognition, and obtains a 97.35\% accuracy on the LFW database. FaceNet~\cite{Schroff2015FaceNet} adopts GoogleNet~\cite{szegedy2015going}, assisted by a private large scale dataset, achieving state-of-art performance (99.63\% on LFW) at that time. SphereNet~\cite{SphereFace} adopts a 64-layer ResNet~\cite{he2016deep} network, with a large-margin loss function, achieving 99.42\% accuracy on the LFW database. ArcFace~\cite{deng2019arcface} develops ResNet~\cite{he2016deep} with an IR block and achieves new state-of-art performance on several benchmarks. 

Despite the success of CNNs, we still wonder can Transformer be used in face recognition and whether it is better than ResNet-like CNNs. Since Transformer has shown its excellent performance combined with large scale databases~\cite{dosovitskiy2020image}, while there have been lots of large scale training database in face recognition. It is interesting to observe the performance of combination of Transformer and large scale face training databases. Perhaps Transformer is just the best to challenge the CNNs hegemony over the face recognition task. It is known that, the efficiency bottleneck of Transformer models, is just the key of them, \emph{i.e}., the self-attention mechanism, which incurs a complexity of $O(n^2)$ with respect to sequence length~\cite{han2020survey}. Of course efficiency is important for face recognition models, but in this paper, let us mainly determine the feasibility of applying Transformer models in face recognition and leave out the potential efficiency problem of them. 

We first experiment with a standard Transformer~\cite{Vaswani17} as ViT~\cite{dosovitskiy2020image} did. However, the original ViT directly flatten the images into patches, which may neglect inter-patch information. Since some of important facial features will be partitioned into different tokens. To better describe the inter-patch information, we slightly modify the tokens generation method of ViT, to make the image patch overlaps, which can improve the performance compared with original ViT and will not increase the computing cost. Face Transformer models are trained on a large scale training database, MS-Celeb-1M~\cite{guo2016msceleb} database, supervised with CosFace~\cite{Wang2018CosFace}, and evaluated on several face recognition benchmarks including LFW~\cite{LFWTech}, SLLFW~\cite{deng2017fine}, CALFW~\cite{zheng2017CALFW}, CPLFW~\cite{CPLFWTech}, TALFW~\cite{zhong2020towards} CFP-FP~\cite{sengupta2016frontal}, AgeDB-30~\cite{moschoglou2017agedb}, and IJB-C~\cite{maze2018iarpa} databases. Finally, we demonstrate that Transformer models trained on a large-scale database obtain comparable performance as CNN with a similar number of parameters and MACs. In addition, it is reasonable to find the Transformer models attend to the face area as we expected. 

The contribution of our work is that we show the feasibility of Transformer models in face recognition and report promising experiment results. How to further improve the performance and efficiency of Transformer models in face recognition is a promising task for future research.

\begin{figure*}[htbp]
	\center
	\includegraphics[width=0.92\linewidth]{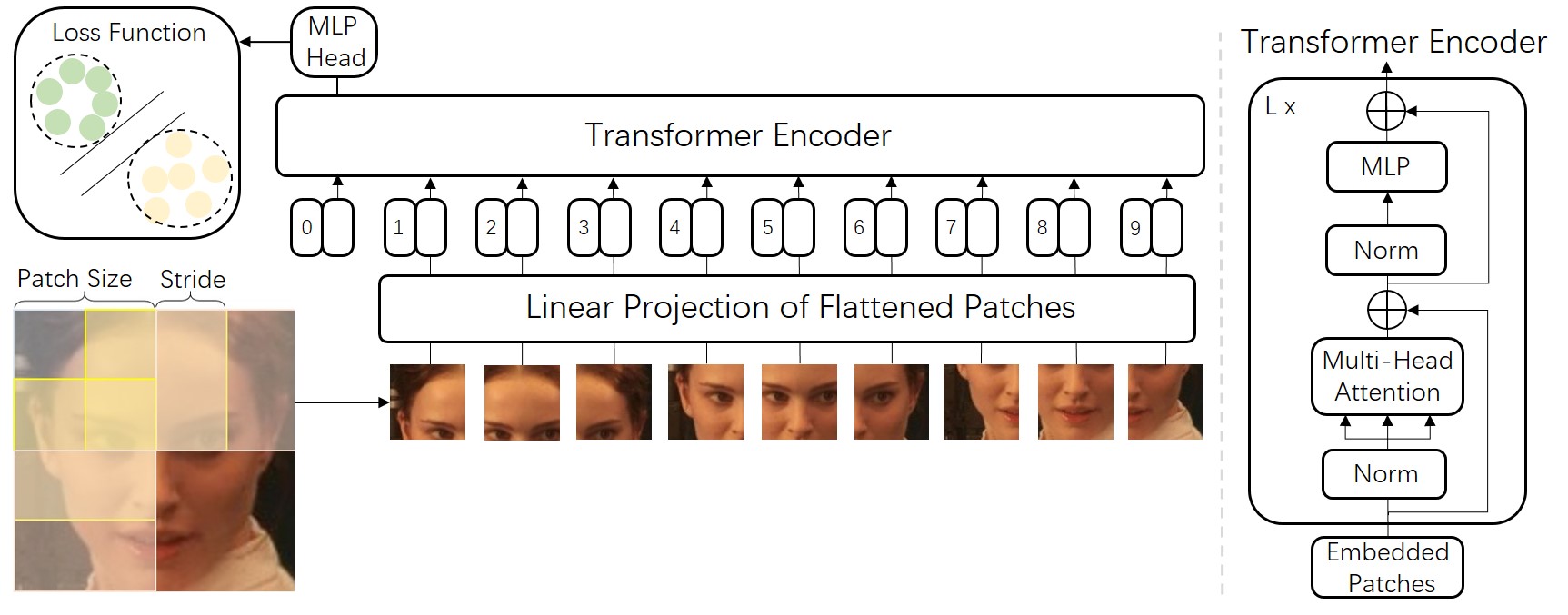}
	\caption{The overall of Face Transformer. The face images are split into multiple patches and input as tokens to the transformer encoder. To better describe the inter-patch information, we modify the tokens generation method of ViT~\cite{dosovitskiy2020image}, to make the image patch overlaps slightly, which can improve the performance compared with original ViT. The Transformer encoder is basically a standard Transformer model~\cite{Vaswani17}. Eventually, the face image embeddings can be used for loss functions~\cite{Wang2018CosFace,deng2019arcface}. The illustration is inspired by ViT~\cite{dosovitskiy2020image}.}
	\label{fig:arch}
\end{figure*}

\section{Face Transformer}
In this paper, following the open-set face recognition pipeline~\cite{SphereFace}, Face Transformer is trained on face databases (with image $\bm{X}$ with label $y$) in a supervised manner, where face images are encoded using a well-designed network, and the output face image embeddings are supervised by an elaborate loss function~\cite{SphereFace,Wang2018CosFace,deng2019arcface} for better discriminative ability, as shown in Figure~\ref{fig:arch}.
 
\subsection{Network Architecture}
Face Transformer model follows the architecture of ViT~\cite{dosovitskiy2020image}, which applies the original Transformer~\cite{Vaswani17}. 

The only difference is that, we modify the tokens generation method of ViT, to generate tokens with sliding patches, \emph{i.e}., to make the image patch overlaps, for the better description of the inter-patch information, as shown in Figure~\ref{fig:arch}. Specifically, we extract sliding patches from the image $\bm{X}\in \mathbb{R}^{W\times W\times C}$ with patch size $P$ and stride $S$ for them (with implicit zero on both sides of input), and finally obtain a sequence of flattened 2D patches $\bm{X_p}\in \mathbb{R}^{N\times( P^2\times C)}$. $(W,W)$ is the resolution of the original image while $(P,P)$ is the resolution of each image patch. The effective sequence length is the number of patches $N=\lfloor \frac{W+2\times p-( P-1 )}{S}+1 \rfloor $, where $p$ is the amount of zero-paddings. 

As ViT did, a trainable linear projection maps the flattened patches $\bm{X_p}$ to the model dimension D, and outputs the patch embeddings $\bm{X_pE}$. The class token, \emph{i.e}., a learnable embedding ($\bm{X}_{class} = \bm{z}_0^0$) is concatenated to the patch embeddings, and its state at the output of the Transformer encoder ($\bm{z}_L^0$) is the final face image embedding, as Equation~\ref{equa:patch_Transformer}. Then, position embeddings are added to the patch embeddings to retain positional information. The final embedding  
\begin{equation}
\label{equa:patch_emb}
\bm{z}_0=\left[ \bm{X}_{class};\bm{X}_{p}^{1}\bm{E};\bm{X}_{p}^{2}\bm{E};\ldots ;\bm{X}_{p}^{N}\bm{E} \right] +\bm{E}_{pos},
\end{equation}serves as input to the Transformer,
\begin{equation}
\label{equa:patch_Transformer}
\begin{gathered}
\bm{z}'_l = MSA(LN(\bm{z}_{l-1}))+\bm{z}_{l-1}, l=1,\ldots, L , \hfill \\
\bm{z}_l = MLP(LN(\bm{z}'_l))+\bm{z}'_l, l=1,\ldots, L , \hfill \\
\bm{x} = LN(\bm{z}_L^0), \hfill \\
\end{gathered}
\end{equation}which consists of multiheaded self-attention (MSA) and MLP blocks, with LayerNorm (LN) before each block and residual connections after each block, as shown in Figure~\ref{fig:arch}. In Equation~\ref{equa:patch_Transformer}, the output $\bm{x}$ is the final output of Transformer model.

One of the key block of Transformer, MSA, is composed of $k$ parallel self-attention (SA), 
\begin{equation}
\label{equ:sa}\begin{gathered}
\left[\bm{q},\bm{k},\bm{v} \right] = \bm{z}\bm{U_{qkv}} , \hfill \\ 
SA(\bm{z}) = softmax(\bm{qk}^T/\sqrt{D_h})\bm{v},
\end{gathered}\end{equation}where $\bm{z} \in \mathbb{R}^{(N+1)\times D}$ is an input sequence, $\bm{U_{qkv}}\in \mathbb{R}^{D\times 3D_h}$ is the weight matrix for linear transformation, and $\bm{A}=softmax(\bm{qk}^T/\sqrt{D_h})$ is the attention map. The output of MSA is the concatenation of $k$ attention head outputs \begin{equation}MSA(\bm{z})=[SA_1(\bm{z});SA_2(\bm{z});\ldots;SA_k(\bm{z})]\bm{U}_{msa},\end{equation}where $\bm{U}_{msa} \in \mathbb{R}^{kD_h \times (D+1)}$. 

\subsection{Loss Function}
The output $x$ of Equation~\ref{equa:patch_Transformer}, \emph{i.e}., the final output of Transformer model, is supervised by an elaborate loss function~\cite{SphereFace,Wang2018CosFace,deng2019arcface} for better discriminative ability, 
\begin{equation}
L =  - \log {P_{y}} = -  \log  \frac{{{e^{\bm{{W_{{y}}}}^T\bm{{x}} + \bm{{b_{{y}}}}}}}}{{\sum\nolimits_{j = 1}^C {{e^{\bm{{W_j}}^T\bm{{x}} + \bm{{b_j}}}}} }}.
\end{equation}where $y$ is the label, ${P_{y}}$ is the predicted probability of assigning $\bm{{x}}$ to class $y$, $C$ is the number of identities, ${{\bm{W}_{j}}}$ is the $j$-th column of the weight of the last fully connected layer, and $\bm{b_j}\in {\mathbb{R}^{\text{C}}}$ is the bias. Softmax based loss functions~\cite{normface,SphereFace,Wang2018CosFace,deng2019arcface} remove the bias term and transform $\bm{{W_j}}^T\bm{{x}} = s\cos {\theta_j}$, and incorporate large margin in the $\cos{\theta_{y_i}}$ term~\cite{SphereFace,Wang2018CosFace,deng2019arcface}. Therefore, softmax based loss functions can be formulated as\begin{equation}
\label{equation:osoft}
\begin{gathered}
L =  - \frac{1}{N}\sum_{i=1}^N\log {P_{y_i}} 
\hfill \\ 
\quad = - \frac{1}{N}\sum_{i=1}^N\log \frac{{{e^{sf(\theta _{{y_i}})}}}}{{{e^{sf(\theta _{{y_i}})}} + \sum\nolimits_{j = 1,j \ne {y_i}}^C {{e^{s\cos {\theta _j}}}} }}, 
\end{gathered}\end{equation}where $f(\theta _{{y_i}}) = \cos {\theta _{{y_i}}} - m$ in CosFace~\cite{Wang2018CosFace}.

\section{Experiment}
\subsection{Implementation Details}
We apply two training databases, CASIA-WebFace and MS-Celeb-1M~\cite{guo2016msceleb}. CASIA-WebFace is a sweet training database and contains 0.49M images from 10,575 celebrities, which can be seen as a relatively small-scale database compared with million-scale ones~\cite{guo2016msceleb}. MS-Celeb-1M is a popular large scale training database in face recognition and we use the clean version refined by insightface~\cite{deng2019arcface}, which contains 5.3M images of 93,431 celebrities. We choose CosFace~\cite{Wang2018CosFace} ($s$ = 64 and $m$ = 0.35) as the loss function for better convergence and recognition performance. The face images are aligned to 112 $\times$ 112. The Horizontally flip with a probability of 50\% is used for training data augmentation. 

For comparison, the CNN architecture used in our work is a modified ResNet-100~\cite{he2016deep} proposed in the first version of ArcFace paper~\cite{deng2019arcface}, which uses IR blocks (BN-Conv-BN-PReLU-Conv-BN) and applies the ``BN~\cite{ioffe2015batch}-Dropout~\cite{Srivastava2014Dropout}-FC-BN'' structure to get the final 512-$D$ embedding feature. We also experiment with the recent proposed T2T-ViT~\cite{yuan2021tokens}. The number of parameters, MACs and inference speed (Tesla V100, Intel Xeon E5-2698 v4) of these face recognition models are listed in Table~\ref{table:model}. Details are as follows. For ViT models, the number of layers is 20, the number of heads is 8, hidden size is 512, MLP size is 2048. For the Token-to-Token part of T2T-ViT model, the depth is 2, hidden dim is 64, and MLP size is 512; while for the backbone, the number of layers is 24, the number of heads is 8, hidden size is 512, MLP size is 2048. Note that, the ``ViT-P10S8'' represents the ViT model with $10 \times 10$ patch size, with stride $S=8$, and ``ViT-P8S8'' represents no overlapping between tokens. 

\begin{table}[htbp]
\renewcommand\arraystretch{1.1}
\begin{center}
\setlength{\tabcolsep}{2.9mm}{
\caption{Number of parameters, MACs and Inference Speed of Face Recognition Models.}
\label{table:model}
\scalebox{1.1}{
\begin{tabular}{|c|c|c|c|}
\hline
Models&Params (M)&MACs (G)& Img/Sec \\
\hline\hline
ResNet-100~\cite{he2016deep}&65.1&12.1&41.73\\\hline
ViT-P8S8~\cite{dosovitskiy2020image}&63.2&12.4&41.72\\\hline
T2T-ViT~\cite{yuan2021tokens}&63.5&12.7&38.08\\\hline		
ViT-P10S8&63.3&12.4&44.59\\\hline
ViT-P12S8&63.3&12.4&42.45\\\hline
\end{tabular}
}}
\end{center}

\end{table}

\begin{table*}[htbp]
	\renewcommand\arraystretch{1.1}
	\begin{center}
		\setlength{\tabcolsep}{3.18mm}{
		\caption{Performance on LFW~\cite{LFWTech}, SLLFW~\cite{deng2017fine}, CALFW~\cite{zheng2017CALFW}, CPLFW~\cite{CPLFWTech}, TALFW~\cite{zhong2020towards} CFP-FP~\cite{sengupta2016frontal} and AgeDB-30~\cite{moschoglou2017agedb} databases.}
		\label{table:LFWs}
		\scalebox{1.1}{
			\begin{tabular}{|c|c|c|c|c|c|c|c|c|}
				\hline
				Training Data&Models & LFW  & SLLFW & CALFW   & CPLFW  & TALFW  & CFP-FP  & AgeDB-30 \\ \hline\hline
				\multirow{3}{*}{CASIA-WebFace}&ResNet-100~\cite{he2016deep}
				&99.55&98.65&94.13&90.93&53.17&96.30&95.50 \\ \cline{2-9}
				&ViT-P8S8~\cite{dosovitskiy2020image}&97.32&90.78&86.78&80.78&83.05&86.60&81.48 \\ \cline{2-9}
				&ViT-P12S8&97.42&90.07&87.35&81.60&84.00&85.56&81.48 \\ \hline\hline
				\multirow{5}{*}{MS-Celeb-1M}&ResNet-100~\cite{he2016deep}&99.82&99.67&96.27&93.43&64.88&96.93&98.27 \\ \cline{2-9}
				&ViT-P8S8~\cite{dosovitskiy2020image}&99.83&99.53&95.92&92.55&74.87&96.19&97.82 \\ \cline{2-9}
				&T2T-ViT~\cite{yuan2021tokens}&99.82&99.63&95.85&93.00&71.93&96.59&98.07\\ \cline{2-9}
				&ViT-P10S8&99.77&99.63&95.95&92.93&72.95&96.43&97.83 \\ \cline{2-9}
				&ViT-P12S8&99.80&99.55&96.18&93.08&70.13&96.77&98.05 \\ \hline
				
			\end{tabular}
		}}
	\end{center}
\end{table*}

\begin{table}[htbp]
\renewcommand\arraystretch{1.1}
\begin{center}	
	\setlength{\tabcolsep}{3.3mm}{
	\caption{Comparison of different models trained on MS-Celeb-1M on the IJB-C database~\cite{maze2018iarpa}. }
	\label{table:ijbc}
\scalebox{1.1}{
\begin{tabular}{|c|c|c|c|c|}
\hline
\multirow{2}{*}{Models}& \multicolumn{4}{c|}{Verification 1:1 TAR@FAR} \\ \cline{2-5} 
& 1e-4 & 1e-3 & 1e-2 & 1e-1 \\ \hline\hline
	ResNet-100~\cite{he2016deep}&96.36&97.36&98.41&99.13 \\ \hline
ViT-P8S8~\cite{dosovitskiy2020image}&95.96&97.28&98.22&98.99 \\ \hline
T2T-ViT~\cite{yuan2021tokens}&95.67&97.10&98.14&98.90 \\ \hline
ViT-P10S8&96.06&97.45&98.23&98.96\\ \hline
ViT-P12S8&96.31&97.49&98.38&99.04 \\ \hline
\end{tabular}
}}
\end{center}

\end{table}

We use AdamW~\cite{loshchilov2017decoupled} and cosine learning rate decay following DeiT~\cite{touvron2020training}. The models are trained from scratch without pre-training. With 1 warmup epoch, the initial learning rate is set as 3e-4, and we lower it to 1e-4 when the training accuracy reaches a stable stage (about 20 epochs). 

\subsection{Results on Mainstream Benchmarks}
We mainly report recognition performance of models on several mainstream benchmarks including LFW~\cite{LFWTech}, SLLFW~\cite{deng2017fine}, CALFW~\cite{zheng2017CALFW}, CPLFW~\cite{CPLFWTech}, TALFW~\cite{zhong2020towards} CFP-FP~\cite{sengupta2016frontal}, AgeDB-30~\cite{moschoglou2017agedb}, and IJB-C~\cite{maze2018iarpa} databases. LFW database contains 13,233 face images from 5,749 different identities, which is a classic benchmark for unconstrained face verification. Similar-looking LFW (SLLFW), Cross-Age LFW (CALFW), Cross-Pose LFW (CPLFW) and Transferable Adversarial LFW (TALFW) databases are constructed based on LFW database, to emphasize similar-looking challenges, cross-age challenge and cross-pose challenge, and adversarial robustness of face recognition. CFP-FP database is built for facilitating large pose variation and AgeDB-30 database is a manually collected cross-age database. IJB-C database contains both still images and video frames to address the unconstrained face recognition.

The experimental results are shown in Table~\ref{table:LFWs} and Table~\ref{table:ijbc}. We first find that in Table~\ref{table:LFWs}, Face Transformer models trained on CASIA-WebFace database performs much worse than ResNet-100. Actually, we find that the accuracy of Face Transformer models trained on CASIA-WebFace can reach a high level as ResNet-100, while models cannot generalize well on test databases, which indicates that the scale of CASIA-WebFace may be not enough for Transformer models. 

While things change when we use a much larger training database, MS-Celeb-1M. The performance of Face Transformer models demonstrate promising results on large-scale face training databases. The performance of Face Transformer is competitive compared to the ResNet-100 with similar number of parameters and MACs. Compared with ``ViT-P8S8'', ``ViT-P10S8'' and ``ViT-P12S8'' have better performance, which demonstrates the overlapping patches can help in some degree. T2T-ViT also obtain good performance, while limited computer source, more hyper-parameters for T2T block remains to try. Another interest point is that, Transformer models obtain a little higher accuracy on TALFW database, which is a database with transferable adversarial noise. Since TALFW database is generated using CNNs as surrogate models, it seems that there is no significant specificality with Transformer in terms of adversarial robustness. It is interesting to explore the performance of combination of Face Transformer models and adversarial training.    

\subsection{Discussion}

\subsubsection{Attention Area Analysis}

\begin{figure}[htbp]
	\center
	\includegraphics[width=1\linewidth]{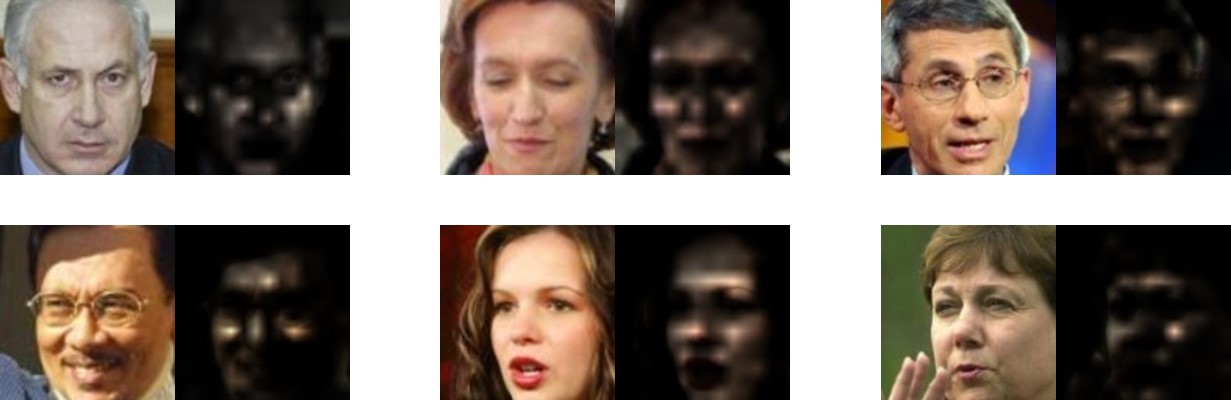}
	\caption{With the help of Attention Rollout~\cite{abnar2020quantifying} techniques, we analyze how the Transformer models (MS-Celeb-1M, ViT-P12S8) concentrate on face images, and find that face Transformer models attend to the face area as we expected.}
	\label{fig:attn_face}
\end{figure}

Since the key of Transformer models is the self-attention mechanism, we analyze how the Transformer models concentrate on face images by analyzing the ViT-P12S8 model trained on MS-Celeb-1M. Specifically, we use the Attention Rollout~\cite{abnar2020quantifying} method, which recursively multiplies the modified attention matrices $0.5\bm{A}+0.5\bm{I}$ of all layers, where $\bm{A}=softmax(\bm{qk}^T/\sqrt{D_h})$ is the attention map of Equation~\ref{equ:sa}. We demonstrate that Transformer models attend to the face area as we expected, as shown in Figure~\ref{fig:attn_face}.

\subsubsection{Attention Matrices Visualization}
\begin{figure}[htbp]
	\center
	\includegraphics[width=0.96\linewidth]{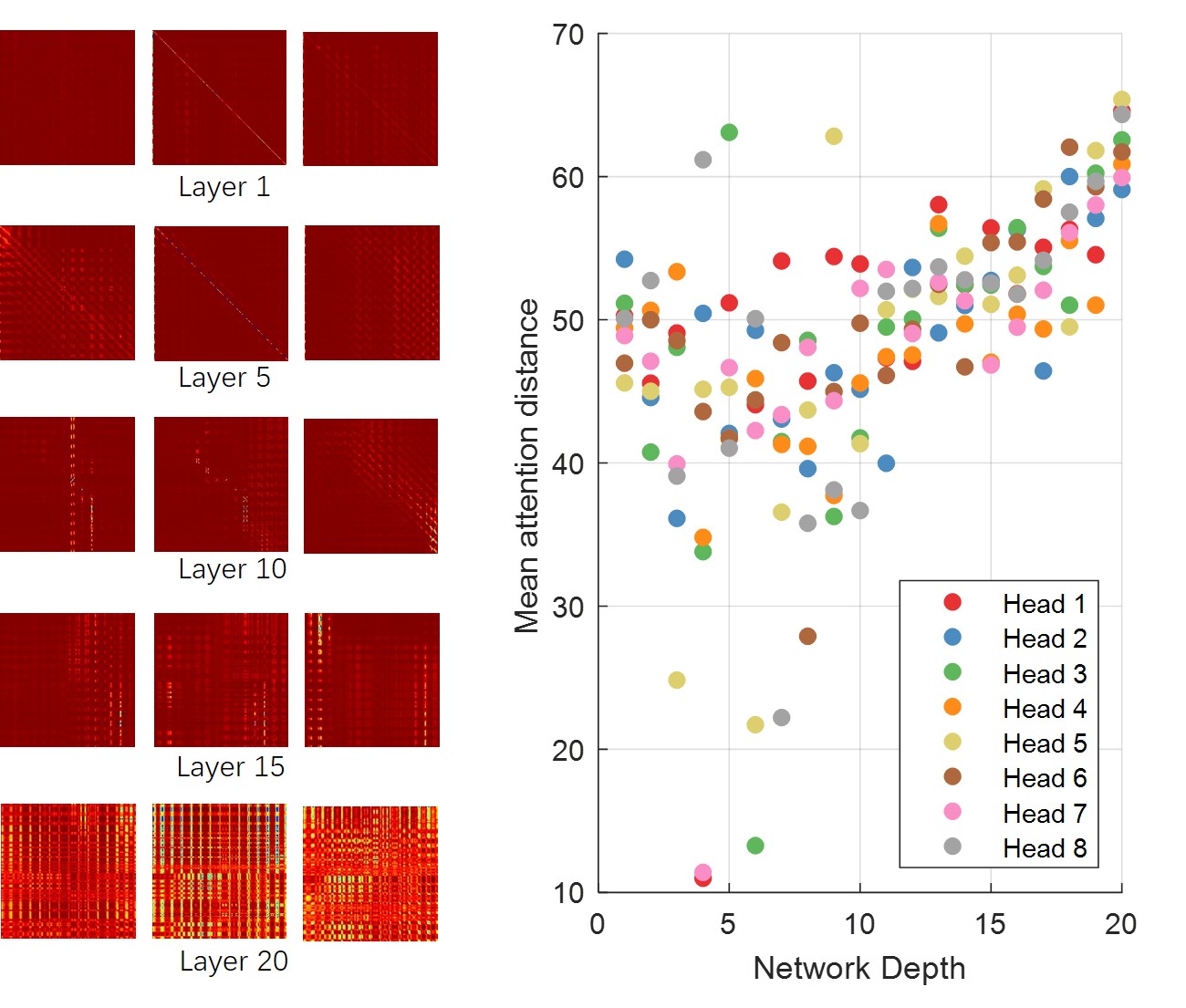}
	\caption{(1) Visualization of the attention matrices of different layers. (2) Mean attention distance of attended area by head and network depth.}
	\label{fig:attn}
\end{figure}

To further understand the Transformer models (MS-Celeb-1M, ViT-P12S8), we visualize the attention matrices of different layers, and calculate the mean attention distance in the image space, which is seemed as the receptive field as CNNs~\cite{dosovitskiy2020image}, shown in Figure~\ref{fig:attn}. While we find that although the deepest layers attend to long distance relationship, it seems that the attention distance of the lowest layer in Face Transformer models is relatively longer than the original ViT~\cite{dosovitskiy2020image}. 

\subsubsection{Occlusion Robustness}
\begin{figure}[htbp]
	\center
	\includegraphics[width=1\linewidth]{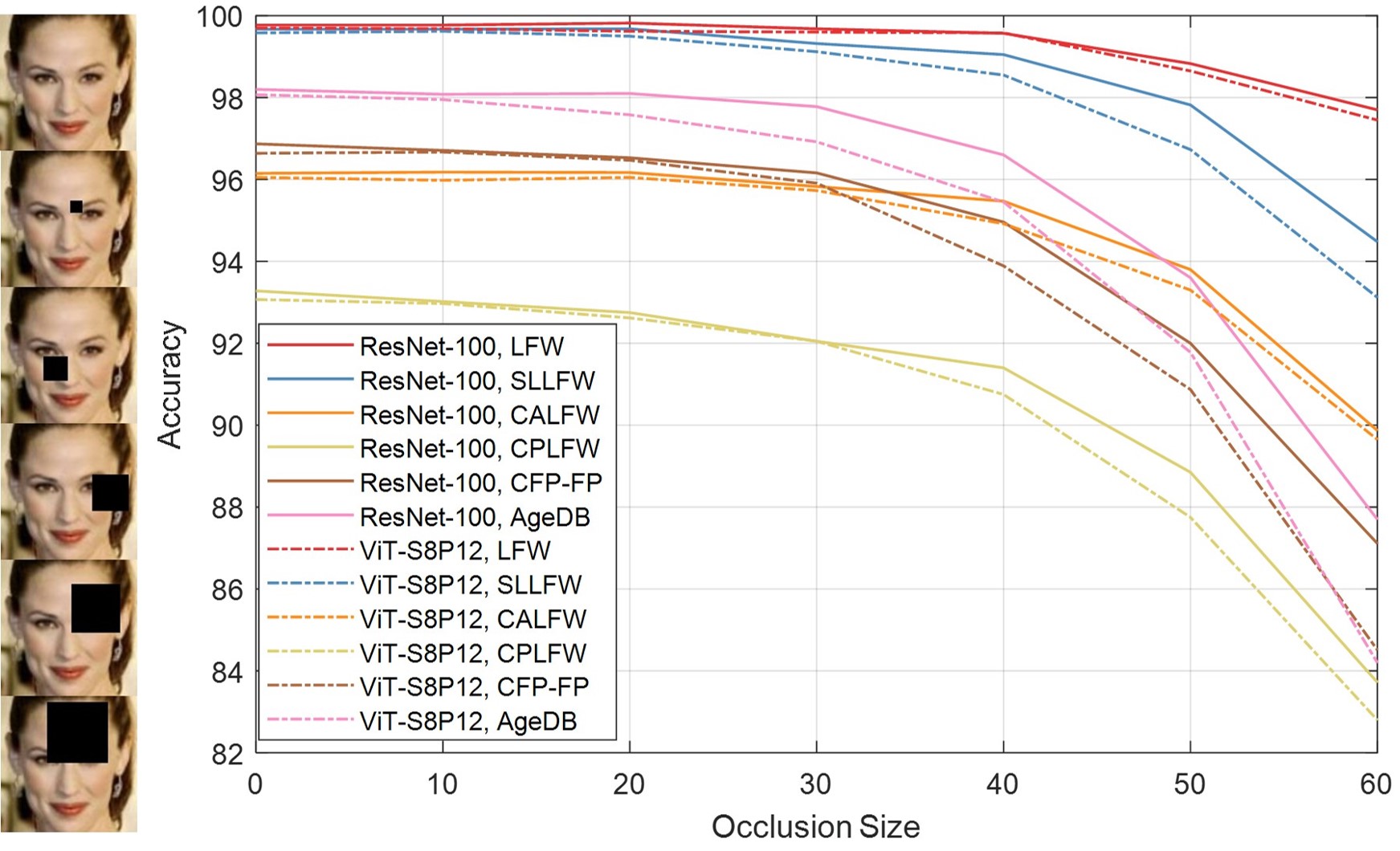}
	\caption{The recognition performance of Face Transformer model and ResNet-100 as the occlusion area increases.}
	\label{fig:occlusion}
\end{figure}

The key of Face Transformer models is the self-attention mechanism and it seems that they concentrate more on the whole face, therefore, we wonder whether them are more robust at classifying partial occluded face images. To explore the occlusion robustness of Face Transformer models, we apply random occlusion (zero value) on face images of several test datasets, and test the recognition performance of models as the occlusion area increases. The experimental results are in Figure~\ref{fig:occlusion}. We find the performance of Face Transformer models decreases more compared with ResNet-100, which indicates Face Transformer models behave no better than CNNs in occlusion robustness. 

\subsubsection{Abortive attempts and Observations}
In addition to the reported models, we would like to share some of our abortive attempts and observations. Note that, these observations may not be rigorous enough to come to conclusions, but maybe they are helpful for readers. 

(1) We first tried SGD as previous works~\cite{Wang2018CosFace, deng2019arcface} to train Face Transformer models, while models cannot converge. So finally we apply AdamW, which has been proved as a effective optimizer for Transformer models.

(2) We tried removing $\bm{X}_{class}(\bm{z}_0^0)$, and used the mean pooling of other tokens outputs. Compared with using $\bm{z}_L^0$ as output, the recognition performance decreases slightly, while accuracy on TALFW database increases to more than 85\%.
 
(3) We tried removing the MLP to improve the efficiency but find the training accuracy cannot increase to a normal level, which indicates that the MLP block is essential for Face Transformer models.

\section{Conclusion}
In this paper, we aim to investigate the feasibility of applying Transformer models in face recognition. Finally, we have demonstrated that Face Transformer models cannot work with a relatively small database, CASIA-WebFace, while they can obtain promising performance on the large-scale face training database, MS-Celeb-1M. In addition, we have provided some analyses for better understanding the Face Transformer models.

\bibliographystyle{IEEEtran}
\bibliography{facetransformer}

\begin{thebibliography}{10}
\providecommand{\url}[1]{#1}
\csname url@samestyle\endcsname
\providecommand{\newblock}{\relax}
\providecommand{\bibinfo}[2]{#2}
\providecommand{\BIBentrySTDinterwordspacing}{\spaceskip=0pt\relax}
\providecommand{\BIBentryALTinterwordstretchfactor}{4}
\providecommand{\BIBentryALTinterwordspacing}{\spaceskip=\fontdimen2\font plus
\BIBentryALTinterwordstretchfactor\fontdimen3\font minus
  \fontdimen4\font\relax}
\providecommand{\BIBforeignlanguage}[2]{{%
\expandafter\ifx\csname l@#1\endcsname\relax
\typeout{** WARNING: IEEEtran.bst: No hyphenation pattern has been}%
\typeout{** loaded for the language `#1'. Using the pattern for}%
\typeout{** the default language instead.}%
\else
\language=\csname l@#1\endcsname
\fi
#2}}
\providecommand{\BIBdecl}{\relax}
\BIBdecl

\bibitem{dosovitskiy2020image}
A.~Dosovitskiy, L.~Beyer, A.~Kolesnikov, D.~Weissenborn, X.~Zhai,
  T.~Unterthiner, M.~Dehghani, M.~Minderer, G.~Heigold, S.~Gelly \emph{et~al.},
  ``An image is worth 16x16 words: Transformers for image recognition at
  scale,'' \emph{arXiv preprint arXiv:2010.11929}, 2020.

\bibitem{carion2020end}
N.~Carion, F.~Massa, G.~Synnaeve, N.~Usunier, A.~Kirillov, and S.~Zagoruyko,
  ``End-to-end object detection with transformers,'' in \emph{European
  Conference on Computer Vision}.\hskip 1em plus 0.5em minus 0.4em\relax
  Springer, 2020, pp. 213--229.

\bibitem{zhou2018end}
L.~Zhou, Y.~Zhou, J.~J. Corso, R.~Socher, and C.~Xiong, ``End-to-end dense
  video captioning with masked transformer,'' in \emph{Proceedings of the IEEE
  Conference on Computer Vision and Pattern Recognition}, 2018, pp. 8739--8748.

\bibitem{touvron2020training}
H.~Touvron, M.~Cord, M.~Douze, F.~Massa, A.~Sablayrolles, and H.~J{\'e}gou,
  ``Training data-efficient image transformers \& distillation through
  attention,'' \emph{arXiv preprint arXiv:2012.12877}, 2020.

\bibitem{yuan2021tokens}
L.~Yuan, Y.~Chen, T.~Wang, W.~Yu, Y.~Shi, F.~E. Tay, J.~Feng, and S.~Yan,
  ``Tokens-to-token vit: Training vision transformers from scratch on
  imagenet,'' \emph{arXiv preprint arXiv:2101.11986}, 2021.

\bibitem{han2021transformer}
K.~Han, A.~Xiao, E.~Wu, J.~Guo, C.~Xu, and Y.~Wang, ``Transformer in
  transformer,'' \emph{arXiv preprint arXiv:2103.00112}, 2021.

\bibitem{guo2016msceleb}
Y.~Guo, L.~Zhang, Y.~Hu, X.~He, and J.~Gao, ``Ms-celeb-1m: A dataset and
  benchmark for large-scale face recognition,'' in \emph{European conference on
  computer vision}.\hskip 1em plus 0.5em minus 0.4em\relax Springer, 2016, pp.
  87--102.

\bibitem{SphereFace}
W.~Liu, Y.~Wen, Z.~Yu, M.~Li, B.~Raj, and L.~Song, ``Sphereface: Deep
  hypersphere embedding for face recognition,'' in \emph{Proceedings of the
  IEEE conference on computer vision and pattern recognition}, 2017, pp.
  212--220.

\bibitem{Wang2018CosFace}
H.~Wang, Y.~Wang, Z.~Zhou, X.~Ji, D.~Gong, J.~Zhou, Z.~Li, and W.~Liu,
  ``Cosface: Large margin cosine loss for deep face recognition,'' in
  \emph{Proceedings of the IEEE Conference on Computer Vision and Pattern
  Recognition}, 2018, pp. 5265--5274.

\bibitem{deng2019arcface}
J.~Deng, J.~Guo, N.~Xue, and S.~Zafeiriou, ``Arcface: Additive angular margin
  loss for deep face recognition,'' in \emph{Proceedings of the IEEE Conference
  on Computer Vision and Pattern Recognition}, 2019, pp. 4690--4699.

\bibitem{simonyan2014very}
K.~Simonyan and A.~Zisserman, ``Very deep convolutional networks for
  large-scale image recognition,'' \emph{arXiv preprint arXiv:1409.1556}, 2014.

\bibitem{he2016deep}
K.~He, X.~Zhang, S.~Ren, and J.~Sun, ``Deep residual learning for image
  recognition,'' in \emph{Proceedings of the IEEE conference on computer vision
  and pattern recognition}, 2016, pp. 770--778.

\bibitem{taigman2014deepface}
Y.~Taigman, M.~Yang, M.~Ranzato, and L.~Wolf, ``Deepface: Closing the gap to
  human-level performance in face verification,'' in \emph{Proceedings of the
  IEEE conference on computer vision and pattern recognition}, 2014, pp.
  1701--1708.

\bibitem{Schroff2015FaceNet}
F.~Schroff, D.~Kalenichenko, and J.~Philbin, ``Facenet: A unified embedding for
  face recognition and clustering,'' in \emph{Proceedings of the IEEE
  conference on computer vision and pattern recognition}, 2015, pp. 815--823.

\bibitem{szegedy2015going}
C.~Szegedy, W.~Liu, Y.~Jia, P.~Sermanet, S.~Reed, D.~Anguelov, D.~Erhan,
  V.~Vanhoucke, and A.~Rabinovich, ``Going deeper with convolutions,'' in
  \emph{Proceedings of the IEEE conference on computer vision and pattern
  recognition}, 2015, pp. 1--9.

\bibitem{han2020survey}
K.~Han, Y.~Wang, H.~Chen, X.~Chen, J.~Guo, Z.~Liu, Y.~Tang, A.~Xiao, C.~Xu,
  Y.~Xu \emph{et~al.}, ``A survey on visual transformer,'' \emph{arXiv preprint
  arXiv:2012.12556}, 2020.

\bibitem{Vaswani17}
A.~Vaswani, N.~Shazeer, N.~Parmar, J.~Uszkoreit, L.~Jones, A.~N. Gomez,
  L.~Kaiser, and I.~Polosukhin, ``Attention is all you need,'' in
  \emph{Advances in Neural Information Processing Systems}, 2017, pp.
  5998--6008.

\bibitem{LFWTech}
G.~B. Huang, M.~Ramesh, T.~Berg, and E.~Learned-Miller, ``Labeled faces in the
  wild: A database for studying face recognition in unconstrained
  environments,'' University of Massachusetts, Amherst, Tech. Rep. 07-49,
  October 2007.

\bibitem{deng2017fine}
W.~Deng, J.~Hu, N.~Zhang, B.~Chen, and J.~Guo, ``Fine-grained face
  verification: Fglfw database, baselines, and human-dcmn partnership,''
  \emph{Pattern Recognition}, vol.~66, pp. 63--73, 2017.

\bibitem{zheng2017CALFW}
T.~Zheng, W.~Deng, and J.~Hu, ``Cross-age {LFW:} {A} database for studying
  cross-age face recognition in unconstrained environments,''
  \emph{arXiv:1708.08197}, 2017.

\bibitem{CPLFWTech}
T.~Zheng and W.~Deng, ``Cross-pose lfw: A database for studying cross-pose face
  recognition in unconstrained environments,'' Beijing University of Posts and
  Telecommunications, Tech. Rep. 18-01, February 2018.

\bibitem{zhong2020towards}
Y.~Zhong and W.~Deng, ``Towards transferable adversarial attack against deep
  face recognition,'' \emph{IEEE Transactions on Information Forensics and
  Security}, vol.~16, pp. 1452--1466, 2020.

\bibitem{sengupta2016frontal}
S.~Sengupta, J.-C. Chen, C.~Castillo, V.~M. Patel, R.~Chellappa, and D.~W.
  Jacobs, ``Frontal to profile face verification in the wild,'' in \emph{2016
  IEEE Winter Conference on Applications of Computer Vision (WACV)}.\hskip 1em
  plus 0.5em minus 0.4em\relax IEEE, 2016, pp. 1--9.

\bibitem{moschoglou2017agedb}
S.~Moschoglou, A.~Papaioannou, C.~Sagonas, J.~Deng, I.~Kotsia, and
  S.~Zafeiriou, ``Agedb: the first manually collected, in-the-wild age
  database,'' in \emph{Proceedings of the IEEE Conference on Computer Vision
  and Pattern Recognition Workshops}, 2017, pp. 51--59.

\bibitem{maze2018iarpa}
B.~Maze, J.~Adams, J.~A. Duncan, N.~Kalka, T.~Miller, C.~Otto, A.~K. Jain,
  W.~T. Niggel, J.~Anderson, J.~Cheney \emph{et~al.}, ``Iarpa janus
  benchmark-c: Face dataset and protocol,'' in \emph{2018 International
  Conference on Biometrics (ICB)}.\hskip 1em plus 0.5em minus 0.4em\relax IEEE,
  2018, pp. 158--165.

\bibitem{normface}
F.~Wang, X.~Xiang, J.~Cheng, and A.~L. Yuille, ``Normface: L2 hypersphere
  embedding for face verification,'' in \emph{Proceedings of the 25th ACM
  international conference on Multimedia}.\hskip 1em plus 0.5em minus
  0.4em\relax ACM, 2017, pp. 1041--1049.

\bibitem{ioffe2015batch}
S.~Ioffe and C.~Szegedy, ``Batch normalization: Accelerating deep network
  training by reducing internal covariate shift,'' \emph{arXiv preprint
  arXiv:1502.03167}, 2015.

\bibitem{Srivastava2014Dropout}
N.~Srivastava, G.~Hinton, A.~Krizhevsky, I.~Sutskever, and R.~Salakhutdinov,
  ``Dropout: a simple way to prevent neural networks from overfitting,''
  \emph{Journal of Machine Learning Research}, vol.~15, no.~1, pp. 1929--1958,
  2014.

\bibitem{loshchilov2017decoupled}
I.~Loshchilov and F.~Hutter, ``Decoupled weight decay regularization,''
  \emph{arXiv preprint arXiv:1711.05101}, 2017.

\bibitem{abnar2020quantifying}
S.~Abnar and W.~Zuidema, ``Quantifying attention flow in transformers,''
  \emph{arXiv preprint arXiv:2005.00928}, 2020.

\end{thebibliography}

\end{document}